\documentclass{article}
\usepackage{spconf,amsmath,epsfig}
\usepackage{multirow}
\usepackage{graphicx}
\usepackage{amsfonts}
\usepackage{bbm}

\let\OLDthebibliography\thebibliography
\renewcommand\thebibliography[1]{
  \OLDthebibliography{#1}
  \setlength{\parskip}{0pt}
  \setlength{\itemsep}{0pt plus 0.3ex}
}

\begin{document}\sloppy

% Example definitions.
% --------------------
\def\x{{\mathbf x}}
\def\L{{\cal L}}

% Title.
% ------
\title{PPGN: PHRASE-GUIDED PROPOSAL GENERATION NETWORK FOR REFERRING EXPRESSION COMPREHENSION}
%
% Single address.
% ---------------

\name{Chao Yang$^{\ast}$, Guoqing Wang$^{\ast}$, Dongsheng Li$^{\dagger}$, Huawei Shen$^{\ddagger}$, Su Feng$^{\ast}$, Bin Jiang$^{\ast}$}
%Address and e-mail should NOT be added in the submission paper. They should be present only in the camera ready paper. 
\address{$^{\ast}$College of Computer Science and Electronic Engineering, Hunan University\\ \{yangchaoedu, wgqbeam, hnufs, jiangbin\}@hnu.edu.cn\\ $^{\dagger}$Microsoft Research Asia. dongsli@microsoft.com\\
	$^{\ddagger}$Institute of Computing Technology, Chinese Academy of Sciences. shenhuawei@ict.ac.cn
	}

\maketitle

\begin{abstract}
Reference expression comprehension (REC) aims to find the location that the phrase refer to in a given image. Proposal generation and proposal representation are two effective techniques in many two-stage REC methods. However, most of the existing works only focus on proposal representation and neglect the importance of proposal generation. As a result, the low-quality proposals generated by these methods become the performance bottleneck in REC tasks. In this paper, we reconsider the problem of proposal generation, and propose a novel phrase-guided proposal generation network (PPGN). The main implementation principle of PPGN is refining visual features with text and generate proposals through regression. Experiments show that our method is effective and achieve SOTA performance in benchmark datasets.
\end{abstract}
\section{Introduction}
\label{sec:intro}
 Reference expression comprehension (REC) acts as one of core tasks in human-machine interaction. REC can be typically formulated as locating the entity involved in an expression (e.g., \textit{ a little girl wearing a pink shirt and holding a red umbrella}) through a bounding box. REC is challenging because it requires not only to understand the fine-grained semantic information of image and natural language, but also to align and associate them for locating the true region.

\begin{figure}[t]
	
	\centering
	\includegraphics[width=0.48\textwidth]{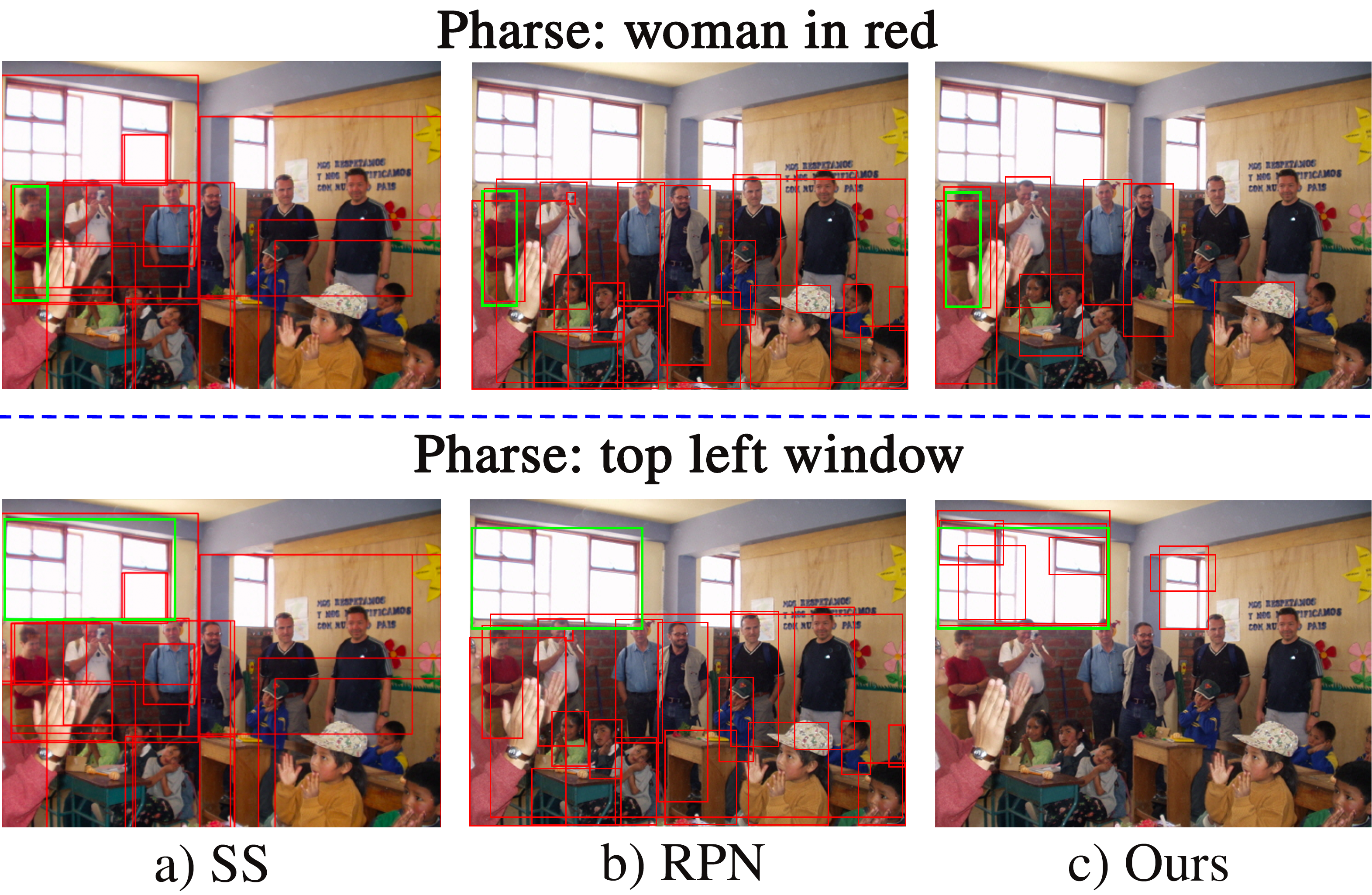}
	\caption{The generated region proposals with different proposal generation methods on COCO. (a) Selective Search (SS); (b) Faster-RCNN pre-trained on COCO;  (c) Our phrase-guided proposal generation network.The red boxes represent the proposals, and the green boxes represent the ground truth.}
	\label{fig:pathdemo4}
\end{figure}
In recent years, mainstream REC works can be categorized into two-stage methods\cite{hu2017modeling,chen2017msrc,wang2019neighbourhood} and one-stage methods\cite{chen2018real,sadhu2019zero,yang2019fast}. Two-stage REC methods were firstly proposed, which can be formulated as follows: Given an input image, a proposal generator is adopted to generate a certain number of region proposals and then the visual features of each proposal are extracted. Simultaneously, a language model (such as Bert\cite{devlin2018bert}) encodes the corresponding referring phrase into language features. The visual features and language features then are fed into the multi-modal fusion module that aims to generate integrated features. After that, the proposal ranking module is utilized to generate the proposal's location with the highest ranking score by using integrated features. Many effective two-stage REC methods have been proposed in recent years, for example, MattNet\cite{yu2018mattnet} parses the phrase into parts of subject, location, and relationship, and links each part with the related object regions for matching score calculation. NMTREE\cite{liu2019learning} parses the phrase via a dependency tree parser and links each tree node with a visual region. DGA\cite{yang2019dynamic} parses the phrase with text self-attention and uses dynamic graph attention to link the text with regions. Nevertheless, recent proposed two-stage REC methods usually focus on the proposal representation and ranking, especially on how to extract more robust visual and text features. In contrast, the problem of proposal generation is rarely explored. Therefore, the resulted proposals are often of low quality that limits the model performance. Thus, in order to avoid the proposal generation stage, the one-stage methods\cite{chen2018real,sadhu2019zero,yang2019fast} have recently been proposed. They fuse visual-text features at the image level and directly predict the boundary box to locate the object they refer to, which significantly improves the model performance and soon becomes prevailing. However, when facing with some complicated scenarios, such as content-rich images or complex semantic expressions, the one-stage methods are sometimes inferior to the two-stage methods\cite{yang2019fast}.

Considering the above problems, it is desirable to pay more attention to proposal generation in two-stage methods. Most existing proposal generation methods can be categorized into non-training methods (e.g., Selective Search\cite{uijlings2013selective}, Edgebox\cite{zitnick2014edge}) and class-aware object detector (e.g., Faster-RCNN\cite{ren2016faster} trained on COCO with $ 80 $ classes). Non-training methods often generate region proposals with some features of the image itself, such as edge information\cite{zitnick2014edge}. As such, proposals generated by these methods have poor discriminability, hence they are difficult to hit the ground truth in a limited number (See Fig. 1(a)). Class-aware object detector can generate proposals discriminatively and accurately only if the referring object pertaining to preset categories\cite{ren2016faster}. However, the vocabulary of referring phrase is usually open, and if the referring object is not in the preset categories, such as “window” is not in the preset categories of COCO, the correct region proposal cannot be generated (See Fig. 1(b)). Moreover, both kinds of methods have a fatal flaw: they generate proposals based only on the image information without considering the referring phrase, which may lead to numerous redundant proposals unrelated to the phrase.

To address the above problems, we propose a novel phrase-guided proposal generation network (PPGN), which is an end-to-end deep regression network. More specifically, PPGN utilizes visual features refined by text features to predict the proposals through regression. In order to make the proposed generation not limited by the preset categories, PPGN is only pretrained on the REC training set with two loss functions that control anchor box offset and confidence respectively. Compared to existing proposal generators, PPGN bears the merits of class-agnostic and high discriminability. Moreover, due to the consideration of phrase information, the generated proposals will change accordingly even for the same image as long as the referring phrase is different (See Fig. 1(c)).

The main contributions of this work are as follows. 1) We propose a novel proposal generation paradigm of REC task, in which the referring phrase directly participates in the proposal generation. 2) We propose a novel phrase-guided proposal generator (PPGN) to generate high-quality proposals. 3) We evaluate our model on the benchmark datasets, and our experimental results show that our model achieves significant performance improvements in the test dataset.

\begin{figure*}[ht]
	
	\centering
	\includegraphics[width=1\textwidth]{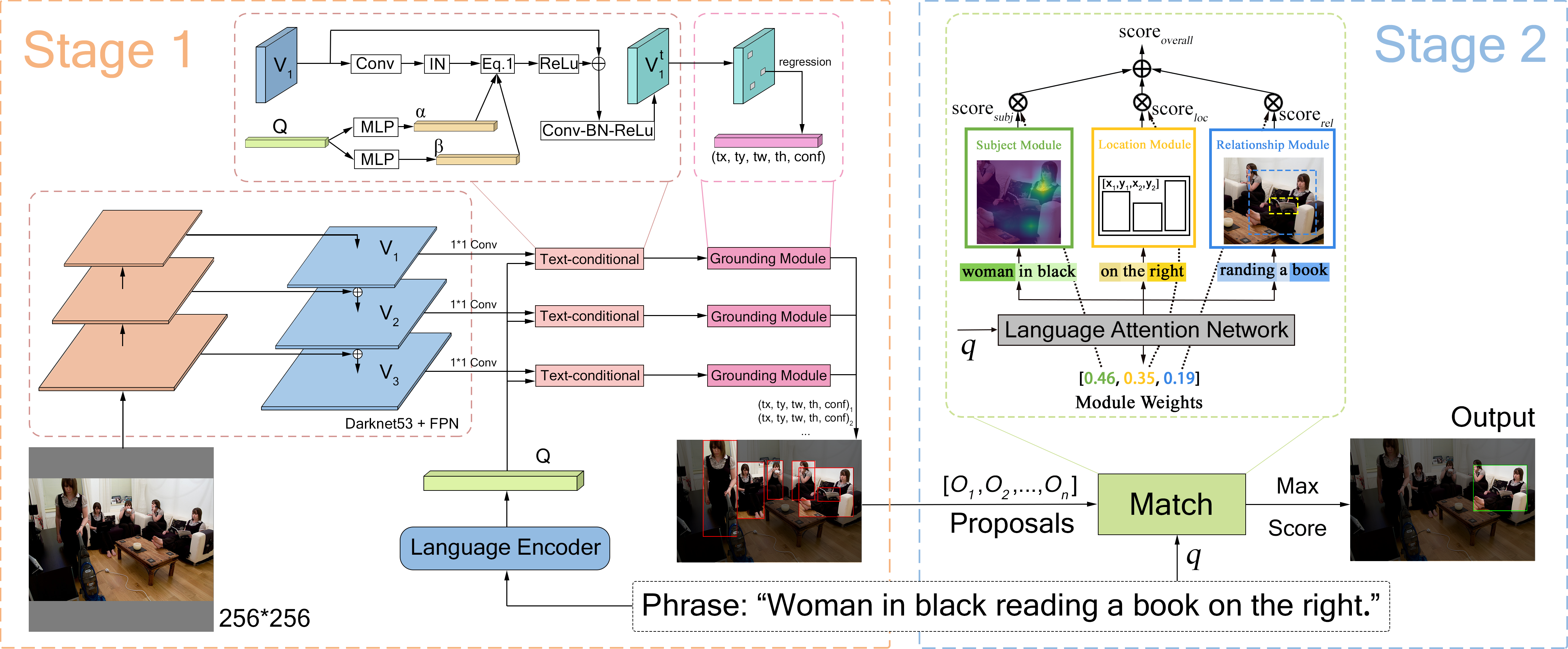}
	\caption{Overall structure of our model. The part of stage 1 is our phrase-guided proposal generation network model, and the part of stage 2 is proposal representation and ranking module. These two parts are combined to make up a complete REC model.}
	\label{fig:pathdemo4}
\end{figure*}

\section{METHOD}
In this section, we will introduce our phrase-guided proposal generation network (PPGN), which is an end-to-end deep regression network. Firstly, we apply feature pyramid network to extract visual feature $ V_{k}(k=1,2,3) $ with three different resolutions of the image, and utilize language model to extract text feature $ Q $ of referring phrase. In order to involve the phrase information in proposal generation, all the visual features are refined with text to obtain the text-conditional visual features $ V^{t}_{k} $ . Then, the model is trained with two loss functions that control anchor box offset and confidence respectively. Finally, the model uses $ V^{t}_{k} $ to predict proposals by regression.
\subsection{Text-conditional Visual Embedding}

 In order to obtain the visual features involving phrase information, we adopt the conditional normalization method in image-level tasks \cite{NIPS2017_6fab6e3a} to refine three different resolution visual features with the text feature. First, text feature $ Q $ is mapped into a scaling vector $ p $ and a shifting vector $ q $ by two MLPs: 
\begin{equation}
 p = tan(W_{p}Q+b_{p}),  q = tan(W_{q}Q+b_{q}), 
\end{equation}
where $ W_{p} $, $ b_{p} $, $ W_{q} $, $ b_{q} $ are learnable parameter. Then, we obtain the text-conditional visual feature $V^{t}_{k}$ via Equation (1) :
\begin{equation}
V_{k}^{t}(i,j) = f_{2}\left\{ReLU[f_{1}(V_{k}(i,j)\odot p+q)+V_{k}(i,j)\right\},
\end{equation}
where $ \odot $ denotes hadamard product, $ (i, j) $ is the spatial coordinate of visual features, $  f_{1} $ and $ f_{2} $ are learnable mapping layers as shown in Fig. 2. $ f_{1} $ is composed of $ 1 \times 1 $ convolution following an instance normalization layer, $ f_{2} $ is composed of a $ 3 \times 3 $ convolution following a batch normalization layer and ReLU activation function. Each coordinate $ (i,j) $ of the visual feature goes through the same operation.

\subsection{Grounding Module }

As shown in the stage 1 of Fig. 2, grounding module takes the text-conditional visual features $ V_{k}(k=1,2,3)  $as input and outputs multiple bounding boxes as the proposals. There are $  8 \times 8+16 \times 16+32 \times 32=1344 $ different locations, and each location corresponds to a vector of a 512-dimensional text-conditional visual feature. Follow YOLOV3\cite{redmon2018yolov3}, in each location, PPGN will set three anchor boxes, and the size of the anchor boxes is obtained by K-means clustering based on the width and height of the ground truth in the training set with (1-IOU) as the distance. Thus, we have a total of $N = 3 \times1344=4032 $ anchor boxes. For each anchor box, the prediction of PPGN is four values though regression for shifting the width, height, and center of the anchor box, together with the fifth value via a sigmoid function about the confidence of the shifted box.

For the prediction of confidence scores, the problem is how to design the loss function to make the predicted confidence scores $ S $ consistent with the ground truth confidence scores $ S^{*} $. Most existing methods\cite{yang2019fast,liu2019improving} define the confidence score $S^{*}=[s_{1}^{*},\ldots ,s_{N}^{*}] \in \left\{ 0,1 \right\} ^{N}$ , which is an one-hot vector that has one only element is set to 1 when the anchor box has the largest IOU with the ground truth box and 0 otherwise. Then they apply the one-hot label to implement the cross entropy loss to train the model.

In this paper, we improve the aforementioned method. Specifically, We set a threshold $ \eta $, calculate the IOU scores between each anchor box and the ground truth box, if it exceeds the threshold $ \eta $, set the IOU score as the label of confidence score, otherwise 0. By doing this, we get the smooth confidence label $ S^{*}=[s_{1}^{*},\ldots ,s_{N}^{*}] \in \mathbb{R}^{N} $, and then perform a L1 regularization to ensure that $ \sum S^{*}=1 $. In this way, we apply the Kullback-Leibler Divergence (KLD) as our loss function to make the smooth confidence label $ S^{*} $ gradually close to the predicted confidence scores $ S $, as shown in Equation(3). Note that $ S $ is also performed L1 regularization to satisfy the same probability distribution.
\begin{equation}
L_{conf}=\dfrac{1}{N}\sum_{n=1}^Ns_{i}^{*}log(\dfrac{s_{i}^{*}}{s_{i}}). 
\end{equation}

The benefits of smooth label are as follows: 1) Some anchors can also provide useful context information besides the anchor with the largest IOU; 2) The soft label actually regularizes the model and alleviates overfitting. 

For the prediction of the the anchor box offset, We adopt the MSE loss function as follows:

\begin{equation}
\begin{aligned}
L_{coord}=&\sum_{n=1}^N \mathbbm{1}^{ IOU}_{n} \Big[(\sigma(t_{x})_{n}-\sigma(\hat{t_{x}})_{n})^{2}\\
& +(\sigma(t_{y})_{n}-\sigma(\hat{t_{y}})_{n})^{2}\\
& +((t_{w})_{n}-(\hat{t_{w}})_{n})^{2}\\
& +((t_{h})_{n}-(\hat{t_{h}})_{n})^{2}\Big],
\end{aligned}
\end{equation}
where $ \mathbbm{1}^{ IOU}_{n} $denotes if the IOU between $ n$-th anchor box and ground truth exceeds the threshold $ \eta $, $ \sigma $ denotes sigmoid function. Note that the loss function only penalizes coordinate error if the anchor box will be selected for a proposal.

The  overall loss of PPGN is defined as:

\begin{equation}
L=L_{conf}+\gamma L_{coord},
\end{equation}
where $ \gamma  $ is a trade-off parameter.

During testing, we first set $K$ to the number of proposals, and then output the shifted anchor boxes in turn according to the confidence of anchor from high to low as proposals. In order to avoid the interference of exception values, we remove the boxes with extremely small height or width. Noted that although our method is similar to recently one-stage REC methods\cite{chen2018real,sadhu2019zero,yang2019fast,liu2019improving}, which also adopt deep regression network, we have a different motivation. One-stage REC methods directly output the only region with the highest confidence as the final result, while our model output multiple regions as proposals.

\begin{table*}[ht]
	
	\caption{Performance Comparison of different REC methods on RefCOCO, RefCOCO+ and RefCOCOg (acc@0.5$\%$). Methods marked with (*) are one-stage methods.}
	\label{tab:my-table}
	\resizebox{\textwidth}{!}
	{%
		\begin{tabular}{l|l|l|ccc|ccc|cc}
			\hline
			\multirow{2}{*}{Method} & \multirow{2}{*}{Proposal Method} & \multirow{2}{*}{Features}& \multicolumn{3}{c|}{RefCOCO} & \multicolumn{3}{c|}{RefCOCO+} & \multicolumn{2}{c}{RefCOCOg} \\
			&                   &                     &   val    & testA      &  testB     &  val    & testA      &  testB    &      val     &  test     \\ \hline
			CMN\cite{hu2017modeling}	&    FRCN Detc.    &    VGG16-COCO        &  -  &  71.03
			& 65.77 & - & 54.32  & 47.76    & -  &- \\
			ParallelAttn\cite{zhuang2018parallel}	&   FRCN Detc.    &    VGG16-ImageNet               &  - &  75.31
			& 65.52 & - & 61.34  & 50.86  & - &- \\
			
			VC\cite{zhang2018grounding}	&    SSD Detec. &   VGG16-COCO
			&  - &  73.33 &67.44 &  - &  58.40
			& 53.18   &  -  &  - \\
			LGRAN\cite{wang2019neighbourhood}	&  FRCN Detc. &    VGG16-ImageNet        &  -  &  76.60
			& 66.40 & - & 64.00  & 53.40  & -  &- \\
			SLR\cite{yu2017joint}	&    SSD Detec.       &    Res101-COCO                &  69.48  &  73.71
			& 64.96 & 55.71 & 60.74  & 48.80   &  60.21  &59.63 \\
			MattNet\cite{yu2018mattnet}	&   FRCN Detc. &  Res101-COCO           & 76.40 & 80.43 & 69.28  & \textbf{64.93 }& 70.26 & 56.00&66.67    &67.01 \\
			DGA\cite{yang2019dynamic}	& FRCN Detc.   &  Res101-COCO &    -  & 78.42      &  65.53  &   -  & 69.07  &51.99  &   -  & 63.28   
			\\
			SSG* \cite{chen2018real}&  -  & Darknet53-COCO &-& 72.51 &67.50   &  - &62.14 & 49.27  & 58.80 &  -\\
			FAOA* \cite{yang2019fast}	&  - & Darknet53-COCO   &72.05&  74.81     & 67.59 & 55.72 & 60.37  &  48.54 &  59.03  & 58.70 \\
			Imp-FAOA*\cite{ yang2020improving}&  - & Darknet53-COCO &  77.63 &80.45 & 72.30 & 63.59 & 68.36& 56.81 & \textbf{67.30} &  67.20\\
			\hline
			Ours-SoftMAX	&   Phrase-guided & Darknet53-COCO   & 77.14      &   80.16    & 72.37  & 64.01 & 69.73 & 57.12 & 66.61
			& 66.29 \\ 
			Ours-KLD	&   Phrase-guided & Darknet53-COCO   & \textbf{77.98}      &   \textbf{81.35}    & \textbf{73.02}  & 64.80 & \textbf{70.42} & \textbf{57.76}  & 67.14
			& \textbf{67.31} \\ \hline
		\end{tabular}%
	}
\end{table*}


\begin{table}[ht]
	\caption{Performance Comparison of different REC methods on Refrit (acc@0.5$ \% $). Methods marked with (*) are one-stage methods.}
	\label{tab:my-table}
	\resizebox{0.48\textwidth }{!}{%
		\begin{tabular}{lllc}
			\hline
			Method& Proposal Method &  Visual Features &Acc\\ \hline
			VC\cite{zhang2018grounding}& SSD Detec. & VGG16-COCO	& 31.13\\
			
			CITE-Resnet\cite{plummer2018conditional}	& Edgebox N=200 &Res101-COCO   & 35.07 \\
			Similarity Net\cite{wang2018learning}	& Edgebox N=200 & Res101-COCO &  34.54 \\
			MattNet\cite{yu2018mattnet}	& FRCN Detc. & Res101-COCO &  29.04 \\
			SSG*	\cite{chen2018real}& - & Darknet53-COCO   & 54.24 \\
			ZSGNet*\cite{sadhu2019zero}	& - & Res50-FPN  &  58.63\\
			FAOA*\cite{yang2019fast}	& - & Darknet53-COCO & 59.30 \\
			Imp-FAOA*	\cite{ yang2020improving}	&-  & Darknet53-COCO  & 64.60 \\ \hline
			Ours-SoftMAX	& Pharse-guided & Darknet53-COCO  & 63.89  \\
			Ours-KLD	& Pharse-guided & Darknet53-COCO  & \textbf{65.65}  \\\hline
		\end{tabular}%
	}
\end{table}

\subsection{Framework Details}

\noindent\textbf{Visual and text feature encoder.} PPGN is an end-to-end network, inputting an image and its corresponding referring phrase, and then outputs a series of image areas as proposals. For the image, we first resize the original image to $ 256\times256 $, and then utilize the Darknet53\cite{redmon2018yolov3} with feature pyramid networks, which is pre-trained on the COCO object detection dataset, to extract visual features. The extracted features have three spatial resolutions, which are $ 8 \times 8 \times D_{1} $, $ 16 \times 16 \times D_{2} $, and $ 32 \times 32 \times D_{3} $. $ D_{1} = 1024 $, $ D_{2} = 512 $, $ D_{3} = 256 $ are the number of feature channels under the corresponding resolution. Finally, we add a 1×1 convolution layer with batch normalization and RELU to map them to the uniform dimension $ D=512 $. For the referring phrase, we embed it to a vector of 768 dimensions via the uncased version of Bert\cite{devlin2018bert}, and then make it through two full connection layers of 512 neurons to get a $ 512D $ text feature. On account of spatial feature encoding will be better achieved in proposal representation and ranking module, we do not repeat this operation in PPGN. 

\noindent\textbf{Proposal representation and ranking module.} To fully implement an REC model, we follow MattNet\cite{yu2018mattnet}, using a modular network to realize the proposal feature representation and ranking. As shown in the stage 2 of Fig. 2, MAttNet applys three modular components related to the appearance, location, and relationship of an object to other objects. The subject module deals with attributes such as categories, colors, and so on. The location module deals with absolute and relative locations, and the relationship module deals with subject-object relationships. Each module has a different structure, learning parameters in its own module space without affecting each other. Instead of using an external language parsers, this module learns to parse phrases automatically by a soft attention mechanism. The matching scores of the three modules are calculated to measure the compatibility between the object and the referring phrase. More details can be found in \cite{yu2018mattnet}. 

\section{EXPERIMENTS}

\subsection{Implementation details}
\textbf{Training setting.} When we resize an input image, we keep the original image ratio and resize its long edge to $ 256 $. We then pad the mean pixel value of the image along the short edge. We adopt the RMSProp optimization method to train the model. We begin with a learning rate of $ 10^{-4} $ and adopt a polynomial schedule with a power of 1. Since Darknet is pre-trained, we reduce the main learning rate of the Darknet portion in the model by 10 times. We set the IOU  threshold $ \eta $ as 0.7, trade-off parameter $ \gamma  $ of loss function as 1. We choose K = 7 as the default number of generated proposals, and relevant ablation experimental studies can be seen in the quantitative results. The batch size is 32 in all of our experiments and we complete our training on a 1080Ti GPU. The training setting of the proposal representation and ranking module is the same as MAttNet\cite{yu2018mattnet}.

\noindent\textbf{Evaluation setting.} We fellow the evaluation method in previous studies\cite{yu2018mattnet,chen2017query,chen2018real,kazemzadeh2014referitgame}, for a given referring phrase, if the IOU between the predict box and the ground truth is not less than 0.5, the predict box is considered correct.

\subsection{Dataset}
We use $ 4 $ classic REC datasets: Refrit\cite{kazemzadeh2014referitgame}, RefCOCO\cite{yu2016modeling}, RefCOCO+\cite{yu2016modeling} and RefCOCOg\cite{mao2016generation}. Refrit contains $ 20,000 $ images from the SAIAPR-12\cite{escalante2010segmented}, and we apply a cleaned split version\cite{chen2017query} with $ 9,000 $, $ 1,000 $, and $ 10,000 $ images in the train, validation, and test sets, respectively. RefCOCO has $ 50,000 $ target entities collected from $ 19,994 $ images. RefCOCO+ has $ 49,856 $ target entities collected from 19,992 images. These two datasets are split into four parts of \textit{train}, \textit{val}, \textit{testA} and \textit{testB}. RefCOCOg includes $ 49,822 $ target entities from 25799 images, which are split into three parts of \textit{train}, \textit{val} and \textit{test}. 

% \usepackage{multirow}
% \usepackage{graphicx}
\begin{table}[t]
	\caption{Comparison of the performance of different proposal generators based on the same proposal representation and ranking module(acc@0.5$ \% $)}
	\label{tab:my-table}
	\resizebox{0.48\textwidth}{!}{%
		\begin{tabular}{l|c|ccc|ccc}
			\hline
			\multirow{2}{*}{Proposal method} &  Referit& \multicolumn{3}{c|}{RefCOCO} & \multicolumn{3}{c}{RefCOCO+}  \\ \cline{2-8} 
			& test &  val   & testA& testB      &  val  &   testA& testB    \\ \hline

			Edgebox\cite{zitnick2014edge}	& 46.54 & 59.32 &  59.09&  56.32  &  50.24  & 55.29 &  45.02 \\
			Selec. Search\cite{uijlings2013selective}&56.45& 56.38   & 55.63 & 54.34 & 51.01 & 53.14  &  42.67  \\
			FRCN Dectc.\cite{ren2016faster}&29.04 &76.40 & 80.43  & 69.28 &\textbf{64.93}  &    70.26   & 56.00 \\ \hline
			Ours& \textbf{65.65} &  \textbf{77.98} &\textbf{81.35} & \textbf{73.02} &  64.80
			& \textbf{70.42}  & \textbf{57.76 } \\ \hline
		\end{tabular}%
	}
\end{table}

\begin{table}[t]
	\caption{Ablation studies on proposal number. (acc@0.5$ \% $)}
	\label{tab:my-able}
	\resizebox{0.48\textwidth}{!}{%
		\begin{tabular}{lcccccccc} 
			\hline
			Num(K)& 1&4 & 7 &10& 13 & 16  \\ 
			\hline
			Acc& 57.25 &64.57&\textbf{65.65}& 64.78 &63.09 & 62.22  \\
			\hline
		\end{tabular}
	}
\end{table}

\subsection{Quantitative Results}

\noindent\textbf{Referring expression comprehension results.}
We perform a comparison of our method with other SOTA methods\cite{hu2017modeling,zhuang2018parallel,zhang2018grounding,wang2019neighbourhood,yu2017joint,yu2018mattnet,yang2019dynamic,chen2018real,yang2019fast,yang2020improving,plummer2018conditional,wang2018learning,sadhu2019zero}. Table. 1 and Table. 2 report the referring expression comprehension results on COCO-series datasets and Referit dataset respectively. The results show that our model outperforms the existing SOTA methods, both two-stage and one-stage. Especially on Referit datasets, which is not a COCO-series dataset, our method exceeds the existing two-stage methods by a large margin. Moreover, the tables also report the performance of our model trained with different losses, which shows that training with the KLD loss result in a 0.6$ \sim $1.8-point improvement over the models with conventional onehot-label softmax loss.

\noindent\textbf{Proposal generator performance comparison.} To further investigate the performance of PPGN, we utilize the same proposal representation and ranking method applied in MAttnet\cite{yu2018mattnet}, changing only the proposal generator for a comparative experiment. We compare three mainstream proposal generators, namely Edgebox\cite{zitnick2014edge}, Select Search\cite{uijlings2013selective} and FRCN object detector\cite{ren2016faster} trained on COCO, and the final results are shown in Table. 3. We can see that our proposed method is superior to other methods in all datasets. FRCN trained on COCO also shows fine performance on the COCO-series datasets (i.e. RefCOCO, RefCOCO+). For example, FRCN performs as well as PPGN in Val and testA of RefCOCO+, though our model significantly outperforms FRCN on RefCOCO and testB of RefCOCO+. However, the performance of FRCN drops dramatically in the Referit dataset. Since RefCOCO/ RefCOCO+ are subsets of COCO and have shared images and entities, the COCO-trained detector can generate almost perfect region proposals on the COCO-series datasets. When we apply it on other datasets, e.g., Referit datasets, their performance will drop significantly. Nevertheless, PPGN works commendably on all datasets.

\noindent\textbf{Ablation studies.} 
We conducte ablation studies on different proposal generation number $ K $ on the Referit dataset, and the results are shown in Table. 4. We observe that increasing the number of generated proposals no longer resulted in improved accuracy after reaching a certain threshold (for example, $ K \geq 7 $ at Referit). Therefore, in our experiment, we choose $ K = 7 $ as the default value. According to our analysis, since our method generates proposals according to the confidence of anchor from high to low, when $ K=7 $, the hit ratio to ground truth is already pretty high. If $ K $ increases further, the hit ratio increase will not be enough to counteract the negative impact of the proposal redundancy.

\subsection{Visualization}
We visualize the REC results on Referit in Fig. 3. It illustrates that PPGN generates different proposals for the same picture with different referring expressions, and the generated proposals have high discriminability and accuracy. For example, in the first column, under the guidance of the phrase \textit{3nd person black shirt}, PPGN generates proposals about all the people in the image, while under the guidance of the phrase\textit{ glass being held by man in blue shirt}, the model generates proposals about all the glass-like objects in the image.

\section{CONCLUSION}
In this paper, we introduce a new phrase-guided REC task proposal generation paradigm in the first time, aiming to solve the problem of ignoring phrase information during proposal generation. Then, a phrase-guided proposal generation network (PPGN) is proposed to produce high-quality proposals. By utilizing visual features refined by phrase information, PPGN has the advantages of generating class-agnostic and high-discriminability proposals. Meanwhile, we design two loss functions to control anchor box offset and confidence respectively. PPGN alleviates the performance bottleneck caused by low-quality proposals in the conventional two-stage REC methods. Extensive experiments on four banchmark datasets show that our model outperforms other SOTA methods on most evaluation indicators. Based on our work, more efficient two-stage REC methods can be further explored.

\begin{figure*}[ht]
	
	\centering
	\includegraphics[width=1\textwidth]{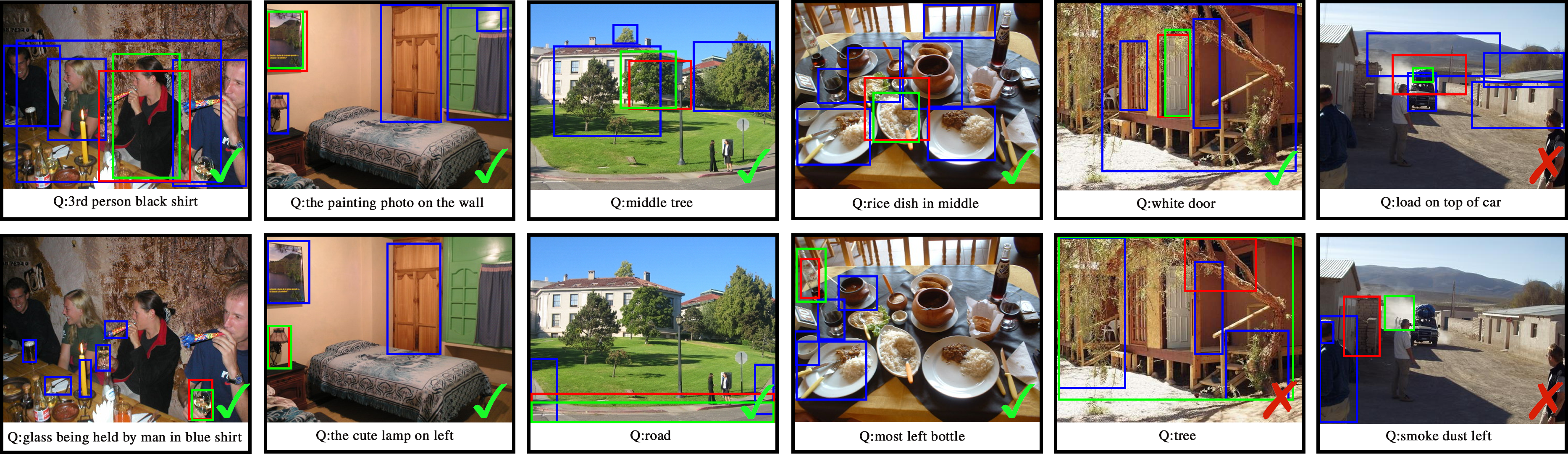}
	\caption{The example in Referit dataset.The ground-truth(green),  the final top-ranked predicted proposal(red), and the other proposals (blue) are visualized respectively. For better viewing, we removed some highly overlapping bounding boxes.We also show 3 examples of incorrect predictions(IoU $ \leq $ 0.5).Best viewed in color.}
	\label{Fig. 3}
\end{figure*}
\bibliographystyle{IEEEbib}
\bibliography{icme2021template}

\end{document}